\documentclass[10pt, conference]{IEEEtran}
\IEEEoverridecommandlockouts
\usepackage{cite}
\usepackage{amsmath,amssymb,amsfonts}
\usepackage{algorithmic}
\usepackage{graphicx}
\usepackage{textcomp}
\usepackage{xcolor}
\usepackage{hyperref}
\usepackage{enumitem}
\usepackage{RobStd}

\usepackage[caption=false,font=footnotesize]{subfig}
\usepackage{siunitx}
\usepackage{tabularx}
\usepackage{booktabs}
\usepackage[letterpaper,margin=1in]{geometry}
\def\BibTeX{{\rm B\kern-.05em{\sc i\kern-.025em b}\kern-.08em
    T\kern-.1667em\lower.7ex\hbox{E}\kern-.125emX}}

\begin{document}

\title{Faithful and Complete LLM-Assisted Automated Construction of Hierarchical Hardware Knowledge Graphs}

\author{
\IEEEauthorblockN{Haotian Qiao and Robert P. Dick}
\IEEEauthorblockA{
Department of Electrical Engineering and Computer Science\\
University of Michigan\\
Ann Arbor, MI, USA\\
qhaotian@umich.edu, dickrp@umich.edu
}
}
\maketitle

\begin{abstract}
Hardware datasheets contain engineering specifications essential for component selection, system integration, and hardware compatibility verification. However, manual inspection is tedious and error-prone because datasheets are lengthy and heterogeneous, with dense information fragmented across tables, figures, and text. Although large language models (LLMs) can automate this process, naïve one-shot processing omits relevant specifications even when the document fits within the model's context window.

We present an automated LLM-driven pipeline that transforms hardware datasheets into general-purpose hierarchical hardware knowledge graphs through chunk-based property extraction, hierarchical routing, and iterative refinement. The framework uses LLMs for information extraction and semantic reasoning while preserving source provenance and applying graph updates through validated instructions and deterministic local execution. This staged design reduces per-query context and maintains traceable intermediate representations. The resulting knowledge graphs provide semantically organized, source-grounded hardware properties for downstream retrieval and reasoning.

We evaluate the framework on 20 hardware datasheets spanning diverse component types and document lengths. We assess property faithfulness and introduce \emph{inter-graph completeness} to measure relative semantic coverage without exhaustive ground truth. Our framework achieves 98.8\% property faithfulness compared with 76.1\% for the one-shot baseline, while fully covering 83.6\% of faithful baseline properties; in the reverse direction, the baseline fully covers only 10.5\% of ours. These improvements provide more reliable and comprehensive engineering information for downstream hardware design and verification, reducing the risk of overlooked constraints and incorrect decisions.
\end{abstract}

\section{Introduction}

Hardware datasheets are a primary source of engineering specifications. They describe electrical characteristics, communication interfaces, operating conditions, functional capabilities, device configurations, and numerous other properties required for downstream engineering tasks such as component compatibility verification and integration. However, the information is typically distributed across lengthy and heterogeneous documents containing narrative descriptions, tables, figures, and device-specific organization. Engineers must therefore repeatedly locate, interpret, and cross-reference specifications from unstructured datasheets, which requires substantial manual effort and causes relevant information to be overlooked. 
For example, datasheets often contain multiple seemingly similar specifications that apply for different operating conditions, variants, or test settings. Designers may locate one value that appears to match the target property and stop searching, even though another specification elsewhere in the datasheet is the one that actually applies to the intended case.

Knowledge graphs (KGs) provide a natural representation for transforming such unstructured documentation into reusable structured engineering knowledge. Rather than representing extracted specifications as an isolated, flat collection of properties, hardware knowledge can be organized according to the functional structure of a device. For example, properties associated with UART, SPI, sensing elements, power supplies, and other hardware functions can be organized by their corresponding functional blocks in a hierarchy. Such a representation gathers related specifications that are scattered throughout the original datasheet and provides a structured search space for downstream applications requiring hardware knowledge retrieval and reasoning. Recent advances in large language models (LLMs) provide an opportunity to automate this transformation through their capabilities in information extraction and semantic reasoning.

\subsection*{Challenges}

One might provide an entire datasheet to an LLM and directly request the resulting knowledge graph. However, this one-shot strategy is poorly suited to reliable hardware knowledge graph construction for three reasons.

First, property omission is frequent in one-shot construction. Hardware datasheets may contain hundreds to thousands of engineering specifications distributed across long and heterogeneous documents, with important information appearing only in isolated table entries or short narrative descriptions. Even when a datasheet fits within the model's context window, prior studies have shown that long-context processing can cause individual pieces of information to be overlooked~\cite{liu2024lost,bai2024longbench,du-etal-2025-context}. This is a major problem for hardware knowledge extraction because engineering importance is not typically correlated with textual prominence: a specification occupying only a small portion of the document may still be critical for downstream design decisions. Consequently, a one-shot approach may produce a plausible-looking KG while omitting relevant engineering properties. Our framework mitigates this problem by partitioning the datasheet into bounded source contexts and restricting each extraction query to a controlled subset of the document, thereby reducing context size and allowing individual specifications to receive more focused and less error-prone analysis by the LLM.

Second, one-shot KG construction requires the LLM to jointly perform several semantically distinct tasks, including property extraction, routing, deduplication, renaming, and KG generation, over the same global context. Prior studies have reported substantial performance degradation on semantic tasks such as named entity recognition and topic extraction across several LLM families~\cite{dimaio2025degradation}, as well as instruction interference when multiple task-specific objectives are combined within a single prompt~\cite{darshan2026gradients}. Our experiments also show that a one-shot formulation that jointly performs these tasks yields substantially lower faithfulness and completeness than our pipeline. However, this degradation cannot be attributed solely to multi-task interference because our pipeline also controls the context and information available to each stage. Moreover, the effects of multi-task inference are task- and model-dependent; prior work has shown that models such as GPT-4 and Llama-2-Chat-70B can benefit from multi-task inference on some composite benchmarks combining classification, question answering, arithmetic, and natural-language inference~\cite{son2024multitask}.

Third, one-shot generation provides limited auditability and control over the KG construction process. Although an LLM can be instructed to report intermediate results, these results are generated within the same inference and therefore cannot be reliably treated as the actual intermediate states from which subsequent outputs are derived. Consequently, when errors occur in the final KG, it is difficult to determine which processing stage introduces them or to evaluate individual stages independently. Moreover, since all tasks operate over the same global context, one-shot generation provides limited control over the information available to each task. In our framework, the output of each stage is explicitly validated and provided as input to the next stage, making the dependencies between successive transformations observable and enforceable while controlling the information available at each stage. This is important in engineering settings because users may need to determine not only whether a property is supported by the datasheet, but also whether it has been correctly organized, merged, renamed, or rerouted before relying on the KG in downstream design tasks. The resulting intermediate representations therefore provide an auditable transformation trace that supports stage-wise inspection, error localization, and evaluation.

We observe that LLMs may unintentionally alter input information when asked to regenerate it in the output, introducing discrepancies into the extracted knowledge. To avoid this, we adopt an instruction-based approach: the LLM generates only explicit operations, which are deterministically executed on the original property data retained on the local computer. This separates semantic reasoning from persistent data manipulation and prevents unintended changes to values, descriptions, conditions, or provenance which may happen in LLM regeneration. Additionally, this separation makes the applied operations explicit and auditable, allowing designers to understand not only what changes were made to the KG but also why they were made, thereby facilitating independent verification of each operation.

To address these challenges, we present an automated LLM-driven framework for constructing general-purpose hierarchical hardware knowledge graphs from embedded-system hardware datasheets. Rather than directly generating a complete KG from an entire datasheet, the framework progressively constructs the graph through chunk-based property extraction, semantic property routing, and iterative refinement. The datasheet is first decomposed into bounded source contexts from which engineering properties are independently extracted while preserving source provenance. Extracted properties are then incrementally organized into a functional hierarchy and refined through deduplication, name conflict resolution, and rerouting (adding blocks and reallocating misplaced properties in the knowledge graph) as the hierarchy evolves.

\subsection*{Problem Definition}
We formulate the hardware knowledge construction problem as follows: given a hardware datasheet, transform its unstructured engineering information into properties (i.e., structured representations of engineering specifications preserving their individual associated engineering contexts) and organize these properties into a hierarchical knowledge graph according to their functional semantics. The graph contains block nodes that form the hardware functional hierarchy and property nodes that represent individual engineering specifications together with their values, units, conditions, configurations, and device variants. Each property is associated with the block to which the specification applies, while shared properties are represented at their highest applicable common functional scope. The resulting KG provides a reusable structured hardware knowledge layer for downstream retrieval and engineering analysis.

\textit{We claim that trustworthy LLM-assisted knowledge graph construction is best viewed not as directly generating a structured knowledge graph from an entire datasheet, but as progressively transforming and organizing engineering information through traceable and controlled stages.} 

Our framework uses LLMs for semantic reasoning while preserving source provenance and executing property and graph transformations through validated instructions and deterministic local programs. By maintaining intermediate properties, source associations, graph structures, and transformation instructions throughout the pipeline, the framework enables transparent, auditable, and source-grounded construction of hardware knowledge graphs. When a system design involves multiple hardware components, our framework constructs a KG for each distinct device or device series based on its corresponding datasheet, allowing the same KG to be reused across components covered by that datasheet.

\subsection*{Contribution}
\begin{tightenumerate}
\item \textbf{Hierarchical hardware specification knowledge extraction and representation:}
We present an automated LLM-driven pipeline that transforms embedded-system hardware datasheets into hierarchical knowledge graphs. The pipeline systematically processes the datasheet, extracts quantitative and qualitative engineering specifications as structured properties, and organizes them according to their hardware functions. This representation organizes related information scattered across the datasheet while preserving its engineering semantics, providing structured knowledge for efficient retrieval and downstream engineering analysis.

\item \textbf{Condition-aware engineering property representation and retrieval:}
Hardware datasheets may specify semantically related properties under different operating conditions, configurations, use cases, or device variants, making isolated property retrieval ambiguous. Our hierarchical representation preserves these applicability conditions and organizes semantically related properties under the most general block nodes to whose descendant block nodes they apply. During retrieval, related specifications and their conditions can therefore be considered together, reducing the risk of overlooking relevant contextual information distributed across the datasheet.

\item \textbf{Automated knowledge graph consistency refinement:}
We introduce an iterative refinement approach that avoids the introduction of duplicated properties, name conflicts, and property misplacement during knowledge graph (KG) construction. By exploiting the hierarchical structure to constrain refinement within semantically related blocks, the framework narrows the search space for deduplication and naming-conflict resolution and reroutes properties as the hierarchy evolves. This produces a cleaner and more consistent KG while substantially reducing LLM context and computational overhead compared with global property comparison.

\item \textbf{Inter-graph completeness measurement and automated evaluation:}
We propose the \emph{inter-graph completeness} measure for relative semantic coverage between knowledge graphs to avoid requiring generally unavailable exhaustive manually labeled ground truth. To reduce granularity variation induced error, properties are first normalized to a consistent semantic granularity; each reference property is then evaluated against the collective information of semantically relevant properties from the other graph while accounting for engineering context. We further automate this normalization, relevance identification, and collective coverage evaluation for scalable KG comparison.

\end{tightenumerate}

\section{Related Works}

ElecGraph-RAG~\cite{anonymous2026elecgraphrag} focuses on reliable retrieval from electronic-component datasheets, avoiding incorrect associations between device entities and extracted attributes. It addresses this problem by constructing device-centric, star-shaped graphs that link normalized device entities to their attributes, and by adding higher-level device profiles and thematic summaries to support coarse-to-fine retrieval. The profiles therefore mainly guide where relevant information should be retrieved from the underlying device--attribute graph.

ChipMind~\cite{xing2026chipmind} addresses multi-round reasoning over long IC design specifications, where a reasoning query may depend on behavioral, logical, conditional, and temporal relationships scattered across multiple sections. It tackles this problem by converting declarative and procedural specification content into structured semantic representations and extracting hierarchical triples that capture entities, actions, conditions, temporal relations, and their dependencies, forming ChipKG for retrieval-augmented reasoning.

These works address different objectives from ours. ElecGraph-RAG primarily focuses on reliable device--attribute association and retrieval, while ChipMind captures inter-property and behavioral dependencies to support multi-round reasoning. In contrast, our work constructs a reusable hardware specification KG that captures how specifications within each device are functionally organized and semantically related. The resulting KG serves as a persistent knowledge representation, organizing quantitative and qualitative engineering properties into functional blocks and sub-blocks while preserving operating conditions, configurations, modes, and device variants as property context. Beyond initial KG construction, our framework further improves the consistency of this representation through localized deduplication, name-conflict resolution, and iterative rerouting of misplaced properties, supporting its reuse across downstream hardware design and verification tasks.

Knowledge graph completeness is an important dimension of KG quality. Issa et al.~\cite{issa2021knowledge} review existing completeness assessment methods and emphasize that completeness evaluation generally depends on a reference or gold standard representing the expected knowledge. Unfortunately, obtaining complete references is often intractable. This challenge is particularly severe for hardware datasheets, where no exhaustive ground-truth set of engineering properties is available. 

A second difficulty is inconsistency in representation granularity: the same engineering knowledge can be grouped or decomposed differently across independently constructed KGs, producing properties with different granularities and making individual properties an unreliable counting unit. For example, one KG may represent several logic-level characteristics within a single property, while another may separately represent $V_{\mathrm{IH}}$, $V_{\mathrm{IL}}$, $V_{\mathrm{OH}}$, and $V_{\mathrm{OL}}$ under the same operating conditions. Therefore, direct property counting or one-to-one property matching can be misleading when corresponding properties have inconsistent granularities, even when the underlying engineering information is equivalent or similar.

To address these challenges, we propose using an \emph{Inter-Graph Completeness} measure between two independently constructed hardware KGs rather than absolute completeness with an impractical ground truth. We first normalize properties to reduce common granularity differences, then perform bidirectional semantic property retrieval and coverage evaluation to determine whether the engineering information represented by each reference property is recoverable from the other KG.

\section{Method}
\begin{figure*}[t]
    \centering
    \includegraphics[width=\textwidth]{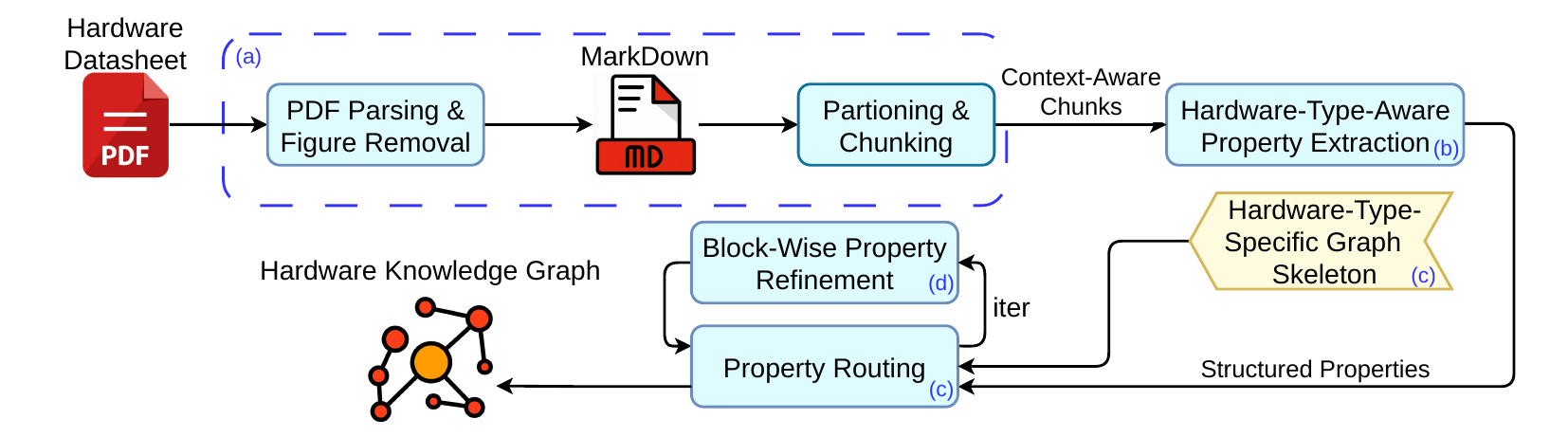}
    \caption{Pipeline of our approach. The PDF is first preprocessed into table-preserved chunks ((a), \autoref{subsec:preprocessing}). Type-aware property extraction is then performed on these chunks ((b), \autoref{subsec:extraction}). Given an initial graph skeleton, the extracted properties are routed to construct the knowledge graph ((c), \autoref{subsec:routing}). Finally, property deduplication, name-conflict resolution, and property rerouting are iteratively applied to refine the hardware knowledge graph until the maximum number of refinement iterations is reached ((d), \autoref{subsec:refinement}).}
    \label{fig:DG-pipeline}
\end{figure*}

\subsection{Document Preprocessing}
\label{subsec:preprocessing}

Hardware datasheets contain dense engineering specifications distributed across prose text, tables, figures, and other document structures. Unlike documents that can often be represented by high-level summaries, datasheets contain numerous individually important details, where even a single table entry or short statement may specify a critical engineering constraint. Processing an entire datasheet in a single LLM query therefore risks omitting fine-grained specifications within a long, information-dense context. We adopt the commonly used chunking strategy to mitigate this problem while preserving datasheet-specific structures. In particular, contextual information, such as table headings and titles, are retained because individual specifications often depend on their row and column contexts. The resulting table-aware chunks serve as the basic input units for subsequent knowledge extraction.

We first convert each PDF datasheet into a structured Markdown representation using MinerU~\cite{wang2024mineru}, preserving textual content and table structures in their original document order. Tables are initially represented in HTML and subsequently converted into a CSV-style representation, which reduces structural markup tokens while preserving the tabular structure and content. Although some hardware specifications are represented through figures, extracting such information requires specialized technical and scientific image understanding, which is beyond the scope of this work. We therefore exclude figures from the Markdown representation.

The resulting Markdown document is partitioned into chunks subject to a maximum token length while preserving sentence boundaries, such that individual sentences are not divided across chunks. Tables are treated differently: each table is retained intact as a single chunk regardless of the token limit. This is necessary because individual table entries often rely on shared headers, titles, captions, units, and condition fields for their engineering meaning. Splitting a table could separate specification values from the contextual information required for their correct interpretation.

Each resulting chunk is independently processed during property extraction. This decomposition bounds the context presented to the LLM in each query and focuses its attention on a portion of the datasheet. Importantly, each extracted property retains its originating chunk as source provenance. This explicit association enables subsequent stages to trace properties directly to their source content and provides source evidence for evaluating whether each extracted property is faithfully grounded in the datasheet.

\subsection{Knowledge Extraction}
\label{subsec:extraction}
Extracting engineering knowledge from chunked hardware datasheets is inherently an open-ended task. Unlike information extraction problems with a predefined static schema, the set of relevant hardware properties cannot be exhaustively specified in advance. Datasheets from different vendors and hardware categories vary substantially in organization, terminology, level of detail, and the types of specifications they provide. Even devices within the same category may expose different functions and characteristics. Consequently, providing deterministic extraction instructions or an exhaustive list of target properties would either restrict extraction to anticipated specifications or require substantial device-specific manual engineering effort, undermining the generality and automation of the framework.

We therefore guide property extraction using high-level engineering descriptions rather than predefined property names or detailed extraction rules. For example, instead of exhaustively enumerating all possible communication interfaces and peripheral functions, an extraction description requests ``supported interfaces (e.g., I2C, SPI, UART, ADC, PWM, GPIO).'' The examples provide semantic guidance to the LLM without restricting extraction to a predetermined set of properties.
The general extraction descriptions identify engineering aspects broadly relevant to embedded-system hardware, including power requirements, communication interfaces, signal characteristics, operating environments, and connector and pin information. These descriptions indicate the types of engineering knowledge of interest while allowing the LLM to identify the specific properties present in each datasheet chunk.

We complement the general descriptions with hardware-category-specific extraction descriptions to capture characteristics relevant to different classes of components. We define category-specific descriptions for four representative embedded-system hardware categories: sensors, processing components, power modules, and communication modules. For example, sensor-specific descriptions describe sensing characteristics and measurement performance, such as measurement precision and accuracy, whereas power-module descriptions emphasize power conversion and regulation characteristics, such as conversion efficiency and dropout voltages. These descriptions remain high-level rather than enumerating individual target properties, allowing the extraction process to accommodate differences among devices and datasheet conventions within the same category.

Each preprocessed chunk is independently provided to the LLM together with the general and corresponding category-specific extraction descriptions. The LLM identifies explicitly stated engineering properties and represents each property using a structured format consisting of a \textit{variable name}, \textit{brief description}, \textit{value or range}, and \textit{unit}. The brief description preserves the engineering context necessary to interpret the property, including applicable operating conditions, configurations, device variants, and test conditions when present. Each extracted property is also associated with its source chunk, preserving provenance.

This decomposition transforms whole-datasheet knowledge extraction into a sequence of modular, bounded-context extraction tasks. Rather than requiring the LLM to search an entire datasheet for an open-ended set of engineering properties in a single query, each query identifies relevant knowledge within a constraint source context, while the high-level extraction descriptions guide the LLM toward engineering information useful for downstream applications.

\subsection{Knowledge Graph Construction}
\label{subsec:routing}
We organize the extracted properties into a hierarchical knowledge graph that captures the functional structure of the hardware and the semantic relationships between hardware functions and their specifications. The graph contains two types of nodes: block nodes and property nodes. A block node represents a hardware component, such as a functional module, interface, subsystem, or other meaningful hardware concept and can contain child block nodes to form a hierarchy; for example, a ``Peripheral'' block may contain ``UART'' and ``SPI'' blocks. A property node represents an individual engineering property extracted from the datasheet and is attached to the block whose function or characteristics it specifies; for example, an ``SPI'' block may contain a property describing its supported CPOL/CPHA modes. Through this organization, the graph represents not only individual engineering specifications but also the functional contexts to which they pertain.

We initialize the graph with a small skeleton defined for each hardware category. The skeleton contains high-level functional blocks commonly shared by devices of the corresponding category and provides the initial structure for property routing. However, a static, fixed hierarchy cannot anticipate the diverse functions and internal structures described across hardware datasheets. We therefore use the skeleton only as a starting point rather than as a predefined static schema that constrains the final graph structure. During subsequent routing procedures, new block nodes are dynamically created and attached to the existing hierarchy according to the actual hardware structure, allowing the graph to accommodate structural variations across different devices.

We define \emph{property routing} as the process of placing an extracted property into the hierarchical knowledge graph based on its engineering semantics and the existing graph structure. For each property, the LLM is provided with the properties to be routed and the current block-node hierarchy, excluding previously routed property nodes to avoid unnecessary context. Based on this information, the LLM identifies the semantically appropriate block node and generates one of two routing operations. 

\begin{enumerate}
\item If an appropriate block already exists, the property is attached under that block. Properties are attached to the most general applicable block in the hierarchy for which they apply unconditionally to the descendant blocks, rather than being redundantly attached to individual descendants. For example, if the minimum output-high voltage (\(V_{\mathrm{OH}}\)) applies generally to peripheral outputs, it is attached to ``Peripheral'' rather than separately to child blocks such as ``UART'' or ``SPI''. In contrast, a UART-specific baud-rate property is attached to ``UART'' because it does not generally characterize other peripheral functions. This routing principle allows shared specifications to be represented once at their common functional scope while retaining more specialized properties under the corresponding lower-level blocks. In this way, the hierarchy grows incrementally from the initial skeleton according to the engineering knowledge encountered in the datasheet.

\item Otherwise, if the property reveals a distinct component or functional concept not represented in the graph, an appropriate block is created and the property is attached to it. Determining whether a new block should be created is inherently open-ended because hardware datasheets vary widely in terminology, organization, and functional decomposition. We therefore do not impose a fixed taxonomy or exhaustive creation rules; instead, the decision to introduce a new block relies on the LLM's semantic reasoning over the property and the current hierarchy. The newly created block is attached under a proper existing block and subsequently becomes available for routing additional properties. For example, given an initial ``Peripheral'' block and a property specifying the maximum supported UART baud rate, the LLM can create a ``UART'' child block under ``Peripheral'' and attach the property to ``UART'', since ``UART'' provides a more specific semantic owner than the general ``Peripheral'' block. 

\end{enumerate}

All routing operations follow an instruction-based execution scheme. The local computer sends queries containing relevant engineering information to the LLM, which returns explicit instructions to reference existing properties and blocks or create new blocks, rather than directly modifying the property set or target KG. The local computer then validates these instructions for structural completeness and consistency with the current graph before executing them through deterministic programs using locally stored data. This separation prevents unintended changes to properties, provenance, or other persistent engineering information unrelated to the intended operation, while still leveraging the LLM's semantic reasoning capability.

This adaptive construction maintains functional-block--property relationships and preserves engineering semantics by representing each property in the context of the hardware function it characterizes rather than as an isolated specification. The resulting hierarchy also provides a structured search space for downstream knowledge retrieval: relevant functional blocks can first be identified, after which property retrieval is restricted to the corresponding portions of the graph. By excluding properties attached to irrelevant blocks, this block-first retrieval substantially reduces the number of properties that need to be considered by the LLM, enabling more focused and compact query contexts for downstream tasks. In contrast, a fully predefined static schema cannot accommodate device-specific functionality, while an unconstrained flat collection of extracted properties fails to capture the functional organization and relationships among specifications.

\subsection{Property Refinement}
\label{subsec:refinement}
The initial property routing produces a hierarchical KG, but independently extracting properties from different datasheet chunks and incrementally constructing the graph may introduce property duplication, naming conflicts, and property misplacement. We therefore introduce an iterative property-refinement process to address these issues while preserving the extracted engineering information.

Property refinement follows the same instruction-based execution scheme as property routing. Properties within the same block are provided to the LLM in batches, and the LLM generates explicit \texttt{MERGE}, \texttt{RENAME}, or \texttt{KEEP} instructions for each input property. A deterministic program running on the local computer executes these instructions on the corresponding KG properties to perform the refinement.

\texttt{MERGE} is defined to identify and consolidate semantically duplicated properties into a single property. \texttt{RENAME} is defined to assign a distinguishing name when identically named properties represent different engineering specifications. \texttt{KEEP} is defined to retain a property without modification. The instructions reference existing properties by their names and unique in-batch indices and are validated before being deterministically executed on the local computer. In particular, the referenced names, indices, and required instruction fields are checked against the input batch and the expected instruction structure.

\begin{figure}[!t]
    \centering
    \includegraphics[width=\columnwidth]{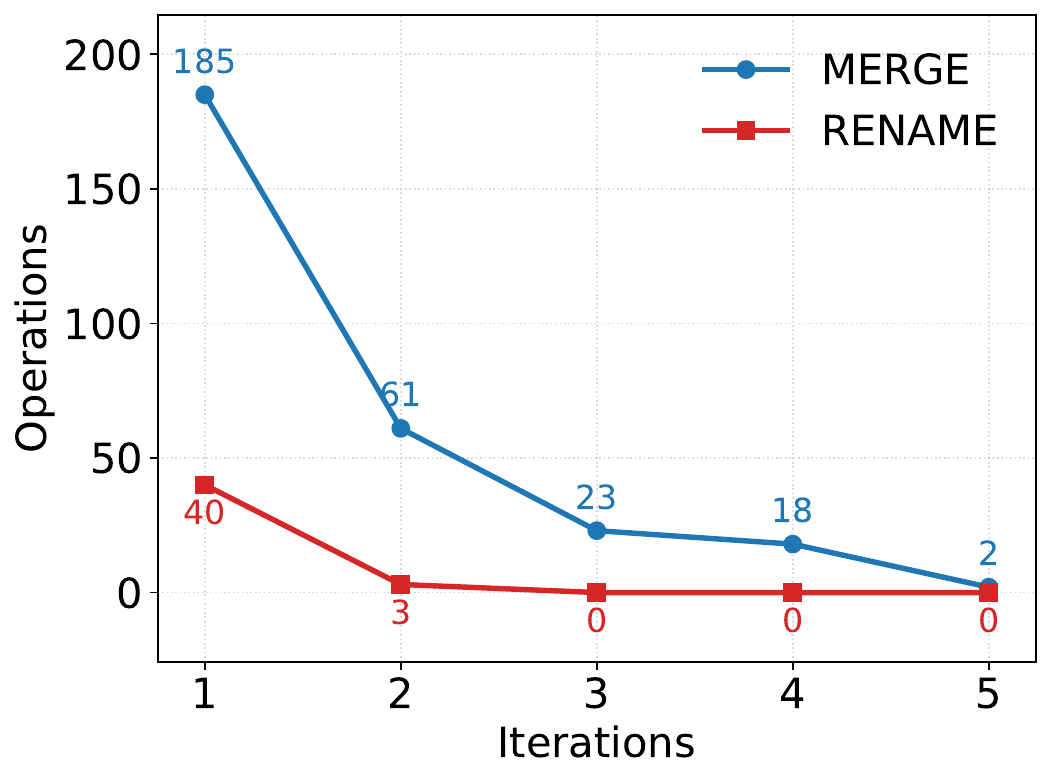}
    \caption{Counts of refinement operations across iterations for the Analog Devices ADXL354/5 sensor. We use three iterations as a practical tradeoff between refinement quality and runtime, as the numbers of \texttt{MERGE} and \texttt{RENAME} operations decrease substantially within the first three iterations. Additional iterations can be configured by users to further reduce remaining duplications at the cost of additional runtime. The refinement process reduces the total number of properties from 1027 to 738.}
    \label{fig:refinement-history}
\end{figure}

We set the maximum number of refinement iterations based on the observed convergence behavior of the refinement process. Our refinement process aims to reach a state in which the LLM identifies no further necessary operations, corresponding to zero additional \texttt{MERGE} and \texttt{RENAME} instructions. Although this criterion does not guarantee that all properties are unique and correctly organized, it provides a practical measure of convergence based on the LLM's semantic judgments. In our experiments, the number of additional operations decreases substantially within three iterations. We therefore adopt three iterations as a practical tradeoff between refinement quality and computational cost, while allowing users to configure additional iterations. 

\autoref{fig:refinement-history} shows an example for the Analog Devices ADXL354/5 sensor, where the number of \texttt{MERGE} operations decreases from 185 in the first iteration to 23 in the third iteration, while \texttt{RENAME} operations decrease from 40 to zero. Although additional iterations can further refine the KG, their marginal benefit becomes relatively small. For the Analog Devices ADXL354/5 sensor KG, which contains more than 800 properties, one complete refinement iteration requires approximately 17 minutes due to the number of LLM queries and associated API and inference latency. This overhead becomes substantial when the framework is applied across more than 30 datasheets, particularly because the Analog Devices ADXL354/5 sensor KG represents a moderately sized KG in our dataset. We therefore use three refinement iterations as a practical tradeoff between refinement quality and runtime. Users may increase the iteration limit when additional refinement is desired and the corresponding computational overhead is acceptable.

\subsubsection{Property Deduplication}

Property duplication occurs when the same engineering specification is described in multiple parts of a datasheet and consequently extracted more than once. Identifying such duplicates globally is challenging because semantically equivalent properties may originate from distant chunks, use different descriptions or representations, and be scattered across a large set of extracted properties. Exhaustively comparing all properties would therefore introduce substantial search and LLM context overhead.

We exploit the hierarchical KG structure to narrow the deduplication scope. Since properties describing the same engineering specification are expected to share the same functional owner, i.e., the same block node, we use each block node as the scope for deduplication and query the LLM to compare properties attached to the same block. This substantially narrows the search space while retaining property groups that are semantically likely to contain duplicates. Importantly, the LLM identifies duplications based on the complete engineering semantics represented by the property fields rather than the variable name alone. Properties with different names can therefore be identified as duplicates if their descriptions, values, units, conditions, and other contextual information collectively represent the same engineering specification. Semantically redundant properties are subsequently consolidated through \texttt{MERGE} instructions. To validate these decisions, we manually inspect 50 \texttt{MERGE} operations. All inspected merges are semantically valid, involving either properties expressing identical engineering specifications or cases in which the information represented by one property is subsumed by the other.

Property deduplication merges only properties that describe the same specification, including cases in which one property is fully subsumed by another. Its goal is therefore not to eliminate all semantic overlap among properties, and the resulting properties are not guaranteed to be mutually exclusive in the information they contain. For example, one property may state that a sensor supports heater power levels of 0, 20, 110, and \qty{200}{mW}, while another states that the maximum supply current is \qty{55}{mA} under the \qty{110}{mW} heater setting. These properties share information about the \qty{110}{mW} configuration but represent different specifications and are not considered duplicates. This distinction is important for inter-graph completeness evaluation: even after deduplication and normalization, a property in one KG may still correspond to multiple properties in the other KG because distinct properties can legitimately contain partially overlapping engineering information without being semantically duplicated, rather than because deduplication or normalization failed.

\subsubsection{Name Conflict Resolution}

name conflicts arise when the same variable name is used for different engineering specifications within the same block. Because the extraction process attempts to preserve the terminology used in the original datasheet, such ambiguity can naturally occur when a datasheet reuses a variable name for multiple specifications. For example, \(V_{\mathrm{DD}}\) is used to describe both an operating supply voltage and an absolute maximum supply voltage in the datasheet of the Espressif ESP32-WROOM-32D/U wireless MCU. If these properties belong to the same functional block, the variable name alone is insufficient to distinguish their engineering meanings.

We therefore resolve name conflicts locally within each block node, consistent with the scope used for deduplication. When identically named properties within a block represent different engineering specifications, the properties remain distinct and \texttt{RENAME} assigns distinguishing names according to their respective engineering semantics. Identical variable names in different blocks are permitted because the unique block index and property name provide an unambiguous identification of semantically distinct properties without requiring globally unique variable names across the KG.

\subsubsection{Property Misplacement}

Property misplacement can arise from the incremental and open-ended nature of graph construction. A property is initially routed to the most appropriate semantic owner available when it is processed. However, a more appropriate block may be created later when another property reveals a distinct hardware or functional concept that was not previously represented. A previously routed property may therefore remain under a more general or less appropriate block even after a better semantic owner becomes available. Because block creation must remain flexible enough to accommodate heterogeneous and previously unseen hardware functionality, the complete hierarchy cannot be deterministically specified in advance.

We solve the property misplacement problem by rerouting properties after deduplication and name-conflict resolution. Unlike the initial routing pass, which starts from the category-specific skeleton, rerouting uses the latest KG structure and exposes only its block nodes to the LLM. Property nodes are intentionally omitted because they are numerous and less necessary for identifying the appropriate target block, thereby reducing context size and focusing the routing decision on the current block hierarchy. The LLM can therefore reconsider property placement with access to blocks that may not have existed when those properties were initially routed. Rerouting otherwise follows the same property-routing procedure described in the previous subsection, allowing properties to be reassigned to more appropriate existing blocks or the creation of new blocks when necessary.

 In summary, through this iterative process described in \autoref{subsec:refinement}, deduplication consolidates redundant properties, name-conflict resolution distinguishes semantically different properties with ambiguous names, and rerouting progressively improves property placement as the hierarchical structure evolves.

\begin{figure*}[!t]
    \centering
    \includegraphics[width=\textwidth]{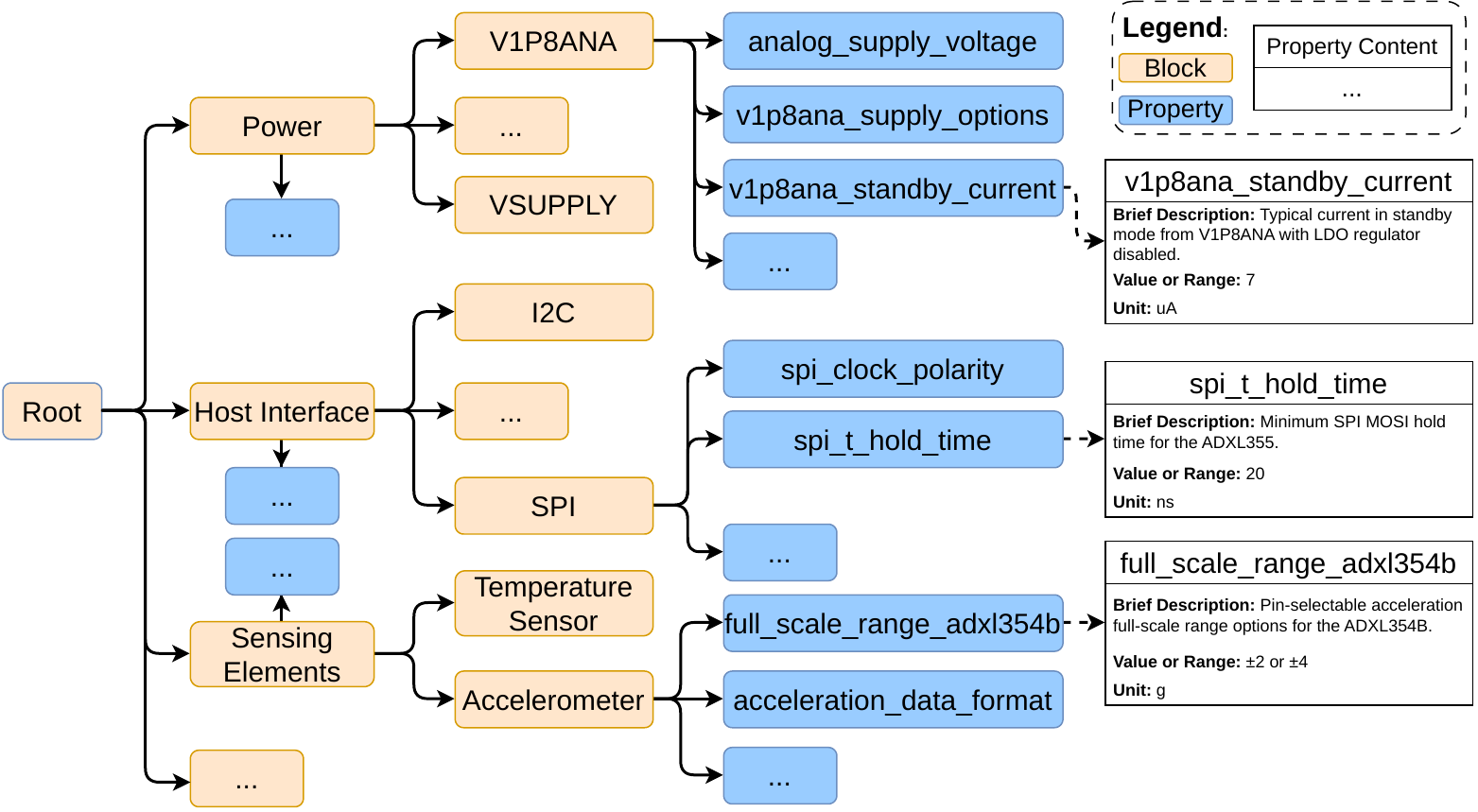}
    \caption{An example knowledge graph for the Analog Devices ADXL354/5 sensor after three refinement iterations. Yellow nodes represent functional blocks and blue nodes represent properties. For clarity, only selected properties are shown; blocks without displayed properties may still have properties attached in the complete KG. The KG organizes engineering properties according to the hardware functional hierarchy. Example property contents are shown as callouts connected to selected property nodes by dashed lines. These callouts are for visualization only and do not represent additional KG nodes; the displayed contents are attributes stored within the corresponding property nodes.}
    \label{fig:kg-example}
\end{figure*}

\autoref{fig:kg-example} shows an example knowledge graph for the Analog Devices ADXL354/5 after three refinement iterations. Yellow nodes represent functional blocks and blue nodes represent properties. For clarity and due to space limitations, only selected properties are shown; blocks without displayed properties may still have properties attached in the complete KG. The resulting KG organizes engineering properties according to the hardware functional hierarchy, with duplicated properties consolidated, properties placed under their appropriate functional blocks, and naming conflicts resolved. This example illustrates how the proposed framework transforms extracted properties into a structured and refined hardware knowledge representation.

\subsection{Knowledge Retrieval and Downstream Applications}

The hierarchical organization of the constructed KG provides a structured search space for downstream applications. Rather than searching all properties globally, knowledge retrieval follows the functional hierarchy of the hardware. Given an engineering query or information requirement, relevant block nodes are first identified according to their functional semantics. Property retrieval is then localized to properties attached to the selected blocks and applicable properties inherited from their ancestor blocks. The retrieved properties can subsequently be filtered or semantically matched according to the requirements of the downstream task. This hierarchical retrieval reduces the property search space while preserving the functional and contextual relationships represented by the KG.

The resulting KGs provide a general-purpose structured hardware knowledge layer that is not limited to a single downstream task. For example, component recommendation and substitution systems~\cite{chipmate2026, similarity2026} could retrieve and compare electrical, functional, and environmental specifications across candidate devices. System-level power estimation tools~\cite{ozkaya2024} could use operating-mode-dependent power specifications to support energy analysis. Similarly, automated hardware design verification systems, such as DRCY~\cite{drcy2026} and DEVICES~\cite{qiao2026llmdrivendatasheetawareautomatedhardware}, could retrieve relevant electrical, interface, power, and environmental properties for compatibility verification. These applications could access source-grounded engineering knowledge through the corresponding functional blocks without repeatedly extracting information from the entire original datasheets.
\section{Evaluation}

\begin{table}[!t]
    \centering
    \caption{Evaluation Dataset Statistics}
    \label{tab:dataset}
    \setlength{\tabcolsep}{6pt}

    \begin{tabular}{lccccc}
        \toprule
        Category & Sensor & Processing & Power & Comm. & Total \\
        \midrule
        Count       & 7   & 5   & 5  & 3   & 20    \\
        Total Pages & 312 & 472 & 85 & 171 & 1,040 \\
        \bottomrule
    \end{tabular}
\end{table}

We evaluate our framework and the baseline using GPT-5.4 with \(temperature=0.2\) on 20 datasheets spanning four categories: sensors, processing components, power modules, and communication transceivers. \autoref{tab:dataset} summarizes the number of datasheets and their total page counts for each category. This section describes the baseline, evaluation measures, property normalization procedure, and analysis of the results.

\subsection{Baseline}

We compare our framework with a one-shot, end-to-end LLM-based KG construction baseline designed to approximate the common practice of providing a complete document and task instructions to an LLM and requesting the resulting knowledge graph. The baseline performs the entire KG construction process in a single query, including identifying relevant engineering properties, preserving their applicable conditions and device variants, deduplicating redundant properties, resolving naming conflicts, and organizing the extracted properties into a hierarchical functional structure similar to that produced by our pipeline, as required by the downstream applications considered in this work. The baseline is also instructed to create new functional block nodes when necessary. The LLM directly generates the final KG as a NetworkX-compatible node-link JSON object containing functional block nodes and extracted engineering properties. For subsequent faithfulness evaluation, we additionally require the LLM to provide the exact supporting text from the input datasheet as a reference for each extracted property.

The baseline prompt uses a straightforward end-to-end task description rather than the detailed, stage-specific instructions of our framework, reflecting typical one-shot LLM document processing while still specifying the necessary construction requirements. These include the target classes of engineering properties, required property fields and contextual information, hierarchical organization, deduplication and naming-conflict handling, and output format. Thus, both approaches share the same overall construction objective, while the baseline must perform extraction, organization, and refinement jointly in a single query.

For a controlled comparison, we provide the baseline with the complete parsed Markdown used by our framework rather than the original PDF. Remote LLM services do not expose their internal PDF parsing or resulting textual representation, making it difficult to determine whether references returned for faithfulness evaluation occur exactly in the processed datasheet. This is particularly important for tabular specifications, whose textual representation depends strongly on how tables are parsed. In addition, direct PDF input would expose figures that are excluded from our framework. Using the same parsed Markdown therefore provides both approaches with identical textual and tabular information, excludes the same figure content, and enables exact string matching of the baseline's returned references against its input.

\subsection{Metrics}
\begin{figure}[!t]
    \centering
    \includegraphics[width=\linewidth]{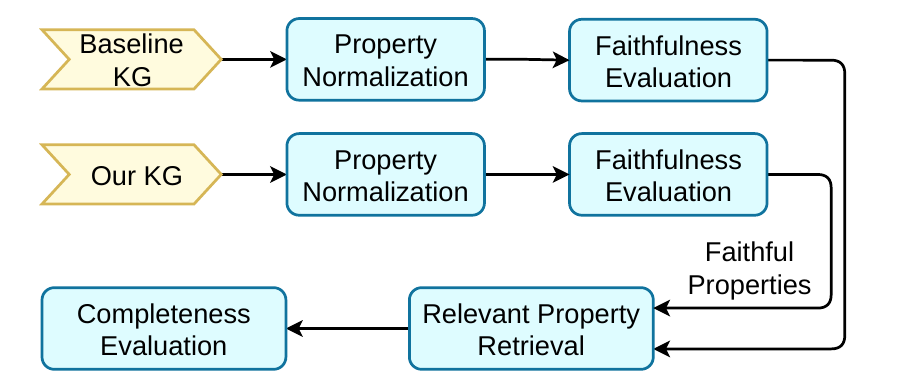}
  \caption{Evaluation pipeline for faithfulness and completeness. Properties from our KG and the baseline are first normalized to reduce granularity differences. Their faithfulness is then evaluated using LLMs, and unfaithful properties are removed. Next, bidirectional relevant-property retrieval is performed: for each property in the Reference KG, all semantically relevant properties are retrieved from the Source KG. Finally, completeness is evaluated for each reference property based on the joint information provided by the retrieved source properties.}
    \label{fig:eval-pipeline}
\end{figure}

\subsubsection{Property Normalization}
\label{subsubsec:property normalization}

Engineering properties can be represented at different granularities. For example, operating conditions may be represented as one property or decomposed into separate temperature, humidity, and pressure properties. Such grouping may be desirable for practical KG usage; for instance, a pressure RMS noise of \(0.2\,\mathrm{Pa}\), equivalent to \(1.7\,\mathrm{cm}\) of altitude variation, can naturally be retained as one property to preserve their semantic relationship. However, granularity differences bias property-level evaluation because the same engineering information may be represented as different numbers of properties to different KGs.

We therefore use an LLM to normalize both KGs to a consistent semantic granularity. We define a normalized property as a single engineering characteristic, capability, requirement, or behavior, together with the context in which it applies or is characterized. Our normalization procedure is rule-based. It is designed to reduce common granularity mismatches while preserving engineering meaning. The LLM applies a fixed set of normalization rules to each property independently. For example, (1) independently specified lower/upper bounds and Min/Typ/Max statistics are split into separate properties, (2) explicitly listed alternatives such as modes, variants, protocols, or supported options are separated into individual assertions, and (3) contextual information, complete equations, and coupled configurations are preserved with the specification to which they apply rather than being split into isolated fragments. Similar rules are used for condition-dependent values and independently bundled specifications, while equivalent representations of the same assertion are retained together. The same rule set is applied uniformly to both KGs.

The goal is not to define a universally canonical property granularity, since engineering information can often be recursively decomposed into increasingly fine-grained but practically meaningless fragments. Instead, the normalization aims to reduce common representation differences consistently across both KGs to enable fair coverage calculation and comparison. The same rule set is applied uniformly to both KGs, and its effectiveness is manually verified.

 Even at a consistent semantic granularity, the same engineering specification may be expressed through multiple properties. For example, ``The main power supply of the Bosch BME280 sensor is 1.71--\qty{3.6}{V}'' may alternatively be represented by two properties: ``VDD is the main power supply pin of BME280'' and ``VDD accepts 1.71--\qty{3.6}{V}.'' Collectively, the two properties convey the same engineering specification despite using a different representation. We therefore account for such collective correspondence in the subsequent completeness evaluation.

After normalization, the total number of properties across the 20 datasheets increases from \qty{12.4}{k} to \qty{17.4}{k}. We manually audited 200 randomly sampled normalization instructions, comprising 50 from each of four datasheets: the Bosch BME280 sensor, Analog Devices MAX30102 sensor, ST STM32F303RE microcontroller, and u-blox MAX-M8 GNSS module. Each instruction was evaluated against its original property and normalized output. Of these, 186 (93\%) correctly followed the normalization rules, 8 (4\%) violated them, and 6 (3\%) were ambiguous because the original wording did not support a definitive judgment. For example, the ST STM32F303RE microcontroller common-mode input voltage range was incorrectly retained as a single property rather than decomposed into separate lower- and upper-bound assertions. The 93\% agreement indicates that the normalization procedure reliably applies the intended decomposition rules in most cases and substantially reduces granularity inconsistencies across the evaluated KGs, although it does not provide error-free canonicalization.

The remaining normalization errors introduce uncertainty into the subsequent completeness evaluation because normalized properties are the counting units. An under-split property may represent multiple canonical assertions as a single evaluation unit, thereby underweighting the corresponding engineering information, whereas an over-split property may represent a single canonical assertion as multiple evaluation units and consequently overweight it. Therefore, even if the subsequent semantic coverage evaluation correctly determines the coverage of each resulting property, normalization errors can bias the aggregate full-, partial-, and no-coverage fractions by altering the number and granularity of the evaluation units. Normalized properties should therefore be viewed as a practical approximation of the ideal canonical assertion units, with the observed 93\% normalization accuracy supporting this reliable approximation while also quantifying an important source of error in the reported completeness results.

\subsubsection{Faithfulness}

We define faithfulness as whether an extracted property is semantically supported by its associated source reference. Importantly, we evaluate faithfulness against the reference rather than searching the entire datasheet for supporting information. Therefore, a property is considered unfaithful if its associated reference does not support it, even if the property may be correct or supported elsewhere in the datasheet. This strict definition measures whether the extracted knowledge is traceable to its claimed source.

For our framework, each property retains the source chunk from which it is extracted, and we therefore use this chunk as its reference. For the one-shot baseline, we use the supporting reference returned by the LLM together with each property. Before evaluating the baseline property, we first verify through exact string matching that its returned reference occurs in the input Markdown; a property whose reference cannot be verified is considered unfaithful.

Following prior work that uses LLMs to evaluate generated information and its factual consistency with source evidence~\cite{min2023factscore, liu2023g,chen2023evaluating}, we use an LLM to determine whether the associated reference semantically supports the engineering information represented by each property. Faithfulness is then measured as the fraction of extracted properties supported by their associated references:
\begin{equation}
    F = \frac{N_{\mathrm{faithful}}}{N_{\mathrm{total}}},
\end{equation}
where \(N_{\mathrm{faithful}}\) is the number of properties supported by valid associated references and \(N_{\mathrm{total}}\) is the total number of extracted properties.

\subsubsection{Completeness}

\paragraph{Ideal Representation and Absolute Completeness}

Different divisions of the same information can exist even when properties have comparable granularity. For example, suppose a sensor supports a low-power mode with a maximum output data rate of \qty{100}{Hz} and a supply current of \qty{10}{\micro\ampere}, and a high-performance mode with a maximum output data rate of \qty{1}{kHz} and a supply current of \qty{200}{\micro\ampere}. This information may be represented as two mode-centric properties: ``the low-power mode supports a maximum output data rate of \qty{100}{Hz} with a supply current of \qty{10}{\micro\ampere}'' and ``the high-performance mode supports a maximum output data rate of \qty{1}{kHz} with a supply current of \qty{200}{\micro\ampere}''. Alternatively, it may be represented as two different properties: ``the maximum output data rate is \qty{100}{Hz} in low-power mode and \qty{1}{kHz} in high-performance mode'' and ``the supply current is \qty{10}{\micro\ampere} in low-power mode and \qty{200}{\micro\ampere} in high-performance mode''. Both representations contain the same engineering information and use two properties of comparable granularity, but place the property boundaries differently.

We therefore first define a structure-independent representation of engineering information. An \emph{assertion} is a canonical atomic predicate representing one independently interpretable engineering claim, independent of the conditions under which the claim applies. Its semantic content is restricted to information that characterizes the structure, capability, behavior, constraint, or quantitative relationships of the engineered object. This includes, for example, set-membership statements such as supported modes or interfaces, equalities and inequalities such as specified values and operating bounds, and functional relationships such as frequency or power efficiency expressions. This characterization is intentionally not exhaustive; rather, it focuses on the kinds of information  in engineering datasheets and specifications, in the way it is presented, i.e., datasheet authors generally provide statements that are tractable to interpret, avoiding those requiring intractable analysis, such as the general program-termination problem captured by the halting problem. The granularity and representation of assertions are imposed by a comprehensive and consistent canonicalization policy applied uniformly across properties.

For example, \emph{SPI mode = 0} is a valid assertion because it expresses a single canonical claim, whereas \emph{SPI mode = 0 and maximum SCK frequency = 10 MHz} is non-atomic because it combines two independently interpretable claims and should therefore be decomposed. Canonicalization may further determine how an otherwise-coherent claim is represented. For example, the range \(1.7 \le V_{DD} \le 3.6\,\mathrm{V}\) is canonically represented by two assertions, \(V_{DD,\min}=1.7\,\mathrm{V}\) and \(V_{DD,\max}=3.6\,\mathrm{V}\).

A \emph{relationship} captures a semantic dependency between assertions, e.g., conditional, modal, variant-specific, or joint-configuration dependencies. Examples include \emph{under}, \emph{when}, \emph{in mode}, \emph{for variant}, or \emph{jointly configured with}. The engineering information in a datasheet can be represented as a canonical assertion--relationship graph ($\mathcal{G}_D$) containing assertions of canonical engineering claims and relationships capturing the semantic dependencies among them:
\begin{equation}
\mathcal{G}_D=(\mathcal{A}_{D},\mathcal{R}_{D})
\end{equation}
where \( \mathcal{A}_D \) is the set of unique canonical assertions (graph nodes) and \(\mathcal{R}_{D}\) is the set of unique semantic relationships among them (edges). The representation is canonical in the sense that semantically equivalent expressions are mapped to the same assertion, while semantically equivalent composite information is decomposed according to the same canonicalization rules. For example, if an SPI configuration is canonically represented by the assertions \(\mathrm{CPOL}=1\) and \(\mathrm{CPHA}=1\), connected by a \emph{joint-configuration} relationship, the semantically equivalent expression ``SPI Mode~3'' does not introduce an additional independent information unit.

This representation avoids over-counting information that is repeated across multiple properties as shared context. For example, suppose KG1 contains three properties specifying the accuracy, sampling frequency, and current under operating mode \(A\), while KG2 contains only one property stating that mode \(A\) is supported. In a property-level comparison, the KG2 property may partially cover all three KG1 properties because each implicitly contains the information that mode \(A\) is supported, resulting in three partial-coverage counts for the same underlying fact. Under the ideal representation, however, KG1 contains a single assertion that mode \(A\) is supported, together with separate assertions for the three specifications and relationships connecting them to mode \(A\). KG2 contains only the assertion that mode \(A\) is supported. The shared information is therefore matched and counted exactly once, rather than once for every property in which it appears as context.

Under this ideal representation, completeness is naturally defined over assertions and relationships rather than properties. Let
\begin{equation}
\mathcal{G}_{K}=(\mathcal{A}_{K},\mathcal{R}_{K})
\end{equation}
denote the engineering information recoverable from KG $K$ after mapping it into the same canonical representation. Assertion and relationship absolute completeness can then be expressed as

\begin{equation}
C_{\mathrm{A}}(K,D)
=
\frac{
|\mathcal{A}_{K}\cap\mathcal{A}_{D}|
}{
|\mathcal{A}_{D}|
},
\end{equation}

and

\begin{equation}
C_{\mathrm{R}}(K,D)
=
\frac{
|\mathcal{R}_{K}\cap\mathcal{R}_{D}|
}{
|\mathcal{R}_{D}|
}.
\end{equation}

Each assertion or relationship contributes to the absolute completeness measures at most once regardless of how many times it is expressed, implied, or reused as context. At this level, coverage is binary for each semantic unit: an assertion or relationship is either represented or not represented.

However, constructing this ideal representation directly is impractical for our evaluation. In addition to the absence of an exhaustive datasheet-level ground truth, properties from both KGs would first need to be transformed into the same globally canonical assertion--relationship representation. 
This requires determining how composite specifications should be decomposed,
resolving whether engineering information should be represented as an assertion or a relationship, and identifying and canonicalizing equivalent assertions and relationships.
These operations require global semantic reasoning over the entire KG and constitute a substantial knowledge-reconstruction problem. Moreover, no generally accepted canonical ontology or decomposition scheme exists for heterogeneous hardware specifications. Consequently, modeling choice errors introduced by this additional complex reconstruction would directly determine the evaluation units and therefore the resulting completeness measurement.

\paragraph{Inter-Graph Completeness}
We use inter-graph completeness measured over normalized properties as a practical approximation of ideal absolute completeness. For each directional evaluation, the KG whose knowledge is being evaluated is defined as the \emph{Reference KG}, while the other KG is defined as the \emph{Source KG}. 

\autoref{fig:eval-pipeline} shows the evaluation pipeline for faithfulness and inter-graph completeness.
Before completeness evaluation, properties classified as unfaithful by the preceding faithfulness evaluation are removed. We then apply the normalization procedure described in \autoref{subsubsec:property normalization} to reduce obvious differences in property granularity. Normalization decomposes compound properties according to consistent rules while preserving the contextual information required to interpret each resulting specification. For each faithful normalized Reference property, an LLM retrieves all semantically relevant Source properties and jointly evaluates whether their information provides \emph{full coverage}, \emph{partial coverage}, or \emph{no coverage}.

Our definition of full coverage is intentionally strict. A Reference property is fully covered only when the retrieved Source properties collectively support its complete engineering assertion, including the quantity or behavior, value, unit, applicable variant, operating conditions, scope, and any relationships required to interpret the specification. Partial coverage indicates that some relevant engineering information is recoverable but at least one required detail is missing, while no coverage indicates that no meaningful portion of the Reference property's engineering information is recoverable.

Joint evaluation of multiple Source properties is necessary because distinct engineering properties can collectively specify or imply another engineering property even after granularity normalization. For example, the BOSCH BME280 sensor datasheet specifies the I\(^2\)C read address/control byte as \(111011X1\), while separately defining the first seven bits, \(111011X\), as the device address and the final bit, \(1\), as the read/write control for a read transaction. Our KG captures the complete read address/control byte as one property, whereas the baseline captures the 7-bit device address and read/write control bit as two distinct properties. These properties are neither duplicates nor alternative decompositions of a single property; they represent distinct engineering assertions that jointly determine the complete read address/control byte. Therefore, the absence of an explicit \(111011X1\) property should not be treated as missing information when the assertion can be fully recovered from the Source properties collectively. This also accounts for legitimate semantic overlap or complementary information among distinct properties that is intentionally preserved by our deduplication procedure.

We perform this evaluation in both directions, alternately treating our KG and the baseline KG as the Reference KG. The resulting full-, partial-, and no-coverage fractions characterize the relative semantic coverage of the two KGs. Inter-graph completeness should therefore be interpreted as a property-level approximation of the ideal assertion--relationship completeness objective rather than an absolute measure of KG completeness.

\subsection{Results}

\begin{figure}[!t]
    \centering
    \includegraphics[width=\columnwidth]{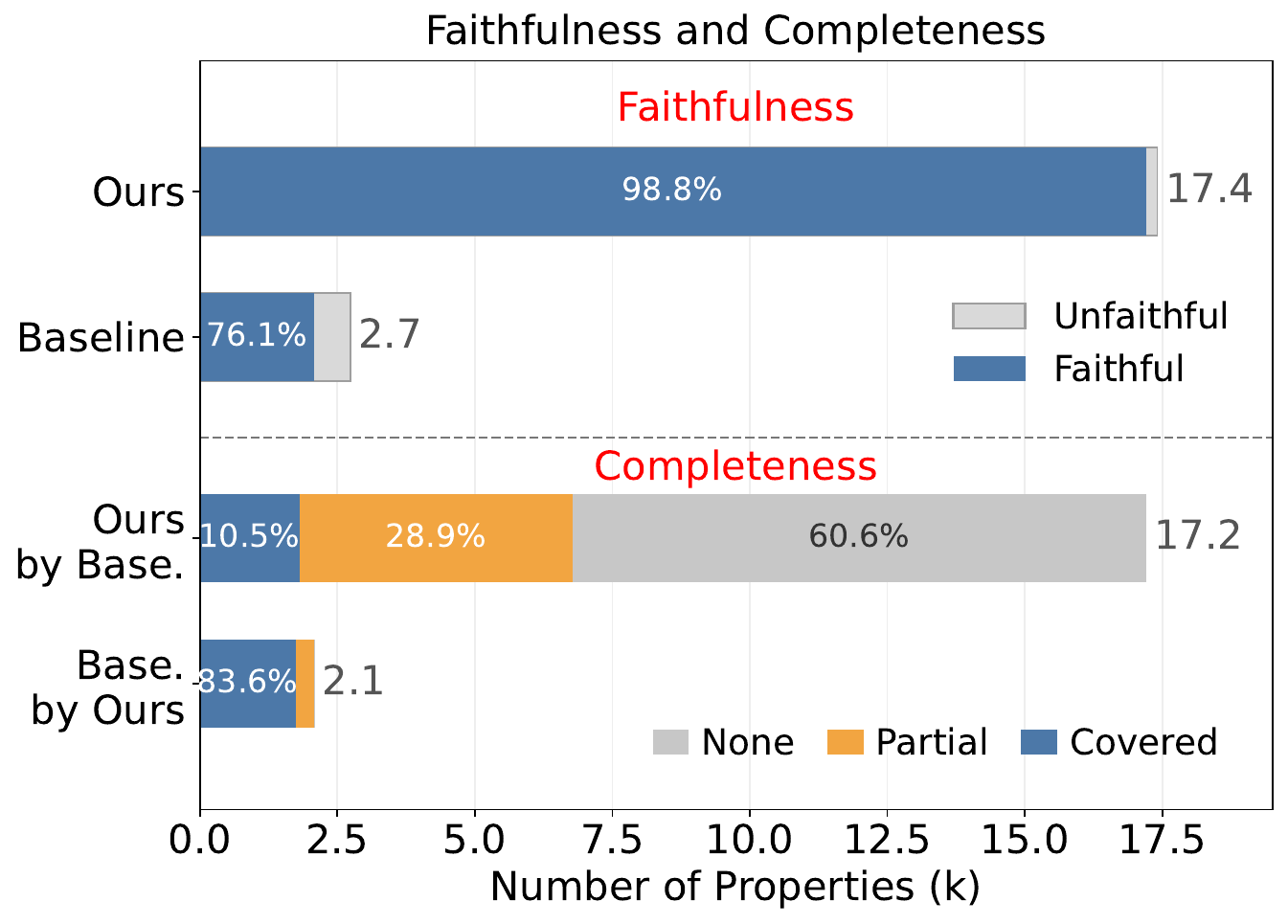}
    \caption{Faithfulness (upper) and completeness (lower) results for our approach and the one-shot baseline across 20 datasheets spanning four hardware categories. Properties in our KGs achieve 98.8\% faithfulness, indicating that nearly all extracted properties are supported by the source datasheets. Our KGs fully cover 83.6\% and partially covers 16.2\% of baseline properties, whereas the baseline fully covers only 10.5\% and partially covers 28.9\% of ours. These results show that, compared with the one-shot baseline, our approach maintains high property faithfulness and constructs KGs with substantially broader engineering information coverage.}
    \label{fig:faithfulness_completeness}
\end{figure}

\autoref{fig:faithfulness_completeness} shows the evaluation results over the dataset. Our KG achieves 98.8\% faithfulness, compared with 76.1\% for the one-shot baseline. Completeness is evaluated only over faithful properties. Our KG fully covers 83.6\% of baseline properties and partially covers the remaining 16.2\%, whereas the baseline fully covers only 10.5\% and partially covers 28.9\% of ours. This strong asymmetry indicates that most faithful information extracted by the baseline is also represented in our KG, while the baseline misses most of the information represented by ours. For downstream applications requiring retrieval or reasoning over datasheet specifications, higher completeness reduces information blind spots: an engineering constraint omitted from the KG cannot be retrieved or considered by subsequent reasoning, regardless of the capability of the downstream model. A more complete KG therefore provides broader coverage of the source engineering information and increases the likelihood that the specifications required by a downstream task are available for consideration.

These results demonstrate that the proposed pipeline produces a substantially more faithful and complete hardware knowledge representation than one-shot KG construction. The high faithfulness indicates that the extracted properties are rarely unsupported by the source datasheets, while the strongly asymmetric inter-graph coverage shows that our KG preserves most of the faithful information captured by the baseline and additionally recovers more engineering information that the baseline omits. Together, the results support the effectiveness of decomposing KG construction into controlled extraction, organization, and refinement stages rather than relying on a single end-to-end LLM generation step.

\subsection{Discussion}

Although the baseline contains only approximately \qty{2.1}{k} faithful normalized properties, it partially covers 28.9\% of our approximately \qty{17.2}{k} properties because partial coverage is not one-to-one. Different reference properties may share contextual or overlapping engineering information, and the same Source property may therefore contribute partial information to multiple Reference properties without fully covering any of them. Our deduplication procedure merges only identical or fully subsumed properties and does not make the information contained in the remaining properties mutually exclusive. Such many-to-many partial coverage is therefore expected.

\section{Conclusions}
This work presents an LLM-assisted framework for constructing hierarchical knowledge graphs from hardware datasheets. Rather than treating KG construction as a one-shot generation task, our framework decomposes the process into explicit stages for property extraction, routing, refinement, and rerouting, while preserving and validating intermediate results throughout the pipeline. The resulting KGs organize engineering properties according to device functional structure while preserving operating conditions, modes, variants, and other applicability contexts required for correct interpretation. An instruction-based execution scheme further separates LLM-generated operations from deterministic manipulation of stored engineering data, improving auditability and reducing unintended modification of extracted properties. Across the evaluated datasheets, our framework achieves 98.8\% faithfulness, while our KGs fully cover 83.6\% of the faithful normalized properties in the one-shot baseline and the baseline fully covers only 10.5\% of ours. These results demonstrate the benefits of decomposed, context-controlled processing for constructing faithful and comprehensive hardware KGs and provide a structured, reusable knowledge representation for downstream hardware design and verification tasks.

\bibliographystyle{IEEEtran}
\bibliography{reference}

@article{issa2021knowledge,
  title     = {Knowledge graph completeness: A systematic literature review},
  author    = {Issa, Subhi and Adekunle, Onaopepo and Hamdi, Fay{\c{c}}al and Cherfi, Samira Si-Said and Dumontier, Michel and Zaveri, Amrapali},
  journal   = {IEEE Access},
  volume    = {9},
  pages     = {31322--31339},
  year      = {2021},
  publisher = {IEEE}
}

@misc{chen2023evaluating,
  title         = {Evaluating factual consistency of summaries with large language models},
  author        = {Chen, Shiqi and Gao, Siyang and He, Junxian},
  year          = {2023},
  eprint        = {2305.14069},
  archivePrefix = {arXiv},
  primaryClass  = {cs.CL},
  url           = {https://arxiv.org/abs/2305.14069}
}

@misc{qiao2026llmdrivendatasheetawareautomatedhardware,
  title         = {{LLM}-driven, datasheet-aware automated hardware compatibility verification for early-stage, pre-schematic embedded system design},
  author        = {Qiao, Haotian and Dick, Robert P.},
  year          = {2026},
  eprint        = {2608.25217},
  archivePrefix = {arXiv},
  primaryClass  = {cs.AI},
  url           = {https://arxiv.org/abs/2608.25217}
}

@article{wang2024mineru,
  title   = {{MinerU}: An open-source solution for precise document content extraction},
  author  = {Wang, Bin and Xu, Chao and Zhao, Xiaomeng and Ouyang, Linke and Wu, Fan and Zhao, Zhiyuan and Xu, Rui and Liu, Kaiwen and Qu, Yuan and Shang, Fukai and others},
  journal = {arXiv preprint arXiv:2409.18839},
  year    = {2024}
}

@inproceedings{bai2024longbench,
  title     = {{LongBench}: A bilingual, multitask benchmark for long context understanding},
  author    = {Bai, Yushi and Lv, Xin and Zhang, Jiajie and Lyu, Hongchang and Tang, Jiankai and Huang, Zhidian and Du, Zhengxiao and Liu, Xiao and Zeng, Aohan and Hou, Lei and others},
  booktitle = {Proceedings of the 62nd Annual Meeting of the Association for Computational Linguistics},
  pages     = {3119--3137},
  year      = {2024},
  publisher = {Association for Computational Linguistics}
}

@article{liu2024lost,
  title   = {Lost in the middle: How language models use long contexts},
  author  = {Liu, Nelson F. and Lin, Kevin and Hewitt, John and Paranjape, Ashwin and Bevilacqua, Michele and Petroni, Fabio and Liang, Percy},
  journal = {Transactions of the Association for Computational Linguistics},
  volume  = {12},
  pages   = {157--173},
  year    = {2024}
}

@inproceedings{xing2026chipmind,
  title     = {{ChipMind}: Retrieval-augmented reasoning for long-context circuit design specifications},
  author    = {Xing, Changwen and Wong, SamZaak and Wan, Xinlai and Lu, Yanfeng and Zhang, Mengli and Ma, Zebin and Qi, Lei and Li, Zhengxiong and Guan, Nan and Jiang, Zhe and others},
  booktitle = {Proceedings of the AAAI Conference on Artificial Intelligence},
  volume    = {40},
  number    = {2},
  pages     = {1337--1345},
  year      = {2026}
}

@inproceedings{liu2023g,
  title     = {{G-Eval}: {NLG} evaluation using {GPT-4} with better human alignment},
  author    = {Liu, Yang and Iter, Dan and Xu, Yichong and Wang, Shuohang and Xu, Ruochen and Zhu, Chenguang},
  editor    = {Bouamor, Houda and Pino, Juan and Bali, Kalika},
  booktitle = {Proceedings of the Conference on Empirical Methods in Natural Language Processing},
  pages     = {2511--2522},
  address   = {Singapore},
  month     = dec,
  year      = {2023},
  publisher = {Association for Computational Linguistics},
  doi       = {10.18653/v1/2023.emnlp-main.153},
  url       = {https://aclanthology.org/2023.emnlp-main.153/}
}

@inproceedings{min2023factscore,
  title     = {{FActScore}: Fine-grained atomic evaluation of factual precision in long form text generation},
  author    = {Min, Sewon and Krishna, Kalpesh and Lyu, Xinxi and Lewis, Mike and Yih, Wen-tau and Koh, Pang and Iyyer, Mohit and Zettlemoyer, Luke and Hajishirzi, Hannaneh},
  editor    = {Bouamor, Houda and Pino, Juan and Bali, Kalika},
  booktitle = {Proceedings of the Conference on Empirical Methods in Natural Language Processing},
  pages     = {12076--12100},
  address   = {Singapore},
  month     = dec,
  year      = {2023},
  publisher = {Association for Computational Linguistics},
  doi       = {10.18653/v1/2023.emnlp-main.741},
  url       = {https://aclanthology.org/2023.emnlp-main.741/}
}

@inproceedings{du-etal-2025-context,
  title     = {Context length alone hurts {LLM} performance despite perfect retrieval},
  author    = {Du, Yufeng and Tian, Minyang and Ronanki, Srikanth and Rongali, Subendhu and Bodapati, Sravan Babu and Galstyan, Aram and Wells, Azton and Schwartz, Roy and Huerta, Eliu A. and Peng, Hao},
  booktitle = {Findings of the Association for Computational Linguistics: EMNLP},
  pages     = {23281--23298},
  year      = {2025},
  publisher = {Association for Computational Linguistics}
}

@inproceedings{anonymous2026elecgraphrag,
  title     = {{ElecGraph-RAG}: A star-shaped graph modeling and efficient hierarchical retrieval framework for electronic datasheets},
  author    = {{Anonymous}},
  booktitle = {Submitted to ACL Rolling Review},
  year      = {2026},
  url       = {https://openreview.net/forum?id=Ywcvhy9JpD},
}

@article{dimaio2025degradation,
  title   = {Degradation of multi-task prompting across six {NLP} tasks and {LLM} families},
  author  = {Di Maio, Federico and Gozzi, Manuel},
  journal = {Electronics},
  volume  = {14},
  number  = {21},
  pages   = {4349},
  year    = {2025},
  doi     = {10.3390/electronics14214349}
}

@inproceedings{darshan2026gradients,
  title     = {When gradients collide: Failure modes of multi-objective prompt optimization for {LLM} judges},
  author    = {Darshan, Parth and Divekar, Abhishek},
  booktitle = {Proceedings of the Second Workshop on Customizable NLP: Progress and Challenges in Customizing NLP for a Domain, Application, Group, or Individual},
  pages     = {216--228},
  year      = {2026},
  publisher = {Association for Computational Linguistics},
  doi       = {10.18653/v1/2026.customnlp4u-1.21}
}

@inproceedings{son2024multitask,
  title     = {Multi-task inference: Can large language models follow multiple instructions at once?},
  author    = {Son, Guijin and Baek, SangWon and Nam, Sangdae and Jeong, Ilgyun and Kim, Seungone},
  booktitle = {Proceedings of the Annual Meeting of the Association for Computational Linguistics},
  volume    = {1},
  pages     = {5606--5627},
  year      = {2024},
  publisher = {Association for Computational Linguistics},
  doi       = {10.18653/v1/2024.acl-long.304}
}

@article{chipmate2026,
  title   = {{ChipMate}: Design-experience-driven chip recommendation framework based on multi-agent collaboration},
  author  = {Xu, Zihan and Liu, Jun and Zhao, Zhenxin and Wang, Yaqi and Wang, Junchao and Zhao, Wen-Sheng},
  journal = {Microelectronics Journal},
  volume  = {174},
  pages   = {107256},
  year    = {2026},
  doi     = {10.1016/j.mejo.2026.107256}
}

@article{similarity2026,
  title   = {Efficient similarity-based datasheet retrieval and analysis using retrieval-augmented generation for electronic component selection},
  author  = {Curavale, Dan and Nicolae, Georgian and Caranica, Alexandru and Cucu, Horia and Burileanu, Corneliu and Davidoiu, Valentina and Buzo, Andi and Pelz, Georg},
  journal = {Electronics},
  volume  = {15},
  number  = {11},
  pages   = {2301},
  year    = {2026},
  doi     = {10.3390/electronics15112301}
}

@article{ozkaya2024,
  title   = {Model-based, fully simulated, system-level power consumption estimation of {IoT} devices},
  author  = {{\"O}zkaya, {\"O}zen and {\"O}rs, Berna},
  journal = {Microprocessors and Microsystems},
  volume  = {105},
  pages   = {105009},
  year    = {2024},
  doi     = {10.1016/j.micpro.2024.105009}
}

@inproceedings{drcy2026,
  title     = {{DRCY}: Agentic hardware design reviews},
  author    = {Dumont, Kyle and Herbert, Nick and Tirmazi, Hayder and Upadhayaya, Shrikanth},
  booktitle = {Proceedings of the ACM Conference on AI and Agentic Systems},
  pages     = {1218--1222},
  year      = {2026},
  publisher = {Association for Computing Machinery},
  doi       = {10.1145/3786335.3813218}
}

\end{document}